\documentclass[runningheads,a4paper]{llncs}

\usepackage{amssymb}
\usepackage{graphicx}
\usepackage{subcaption}
\usepackage{placeins}
\usepackage{color}
\usepackage{fixltx2e}
\usepackage{url}
\usepackage{placeins}
\usepackage[table]{xcolor}
\usepackage{multirow}
\usepackage{booktabs}
\usepackage[hidelinks]{hyperref}
\usepackage{todonotes}
\usepackage[export]{adjustbox}
\usepackage[rightcaption]{sidecap}
\definecolor{red}{rgb}{0.79509804490953684,0.20098039414733648,0.20686274673789687}
\definecolor{blue}{rgb}{0.27900615537575646,0.48814110222692564,0.65689543960433383}
\definecolor{green}{rgb}{0.35022877099759442,0.63449636544374854,0.34172819480299954}
\definecolor{purple}{rgb}{0.56810267475598009,0.34858709455412984,0.591251461777617}
\definecolor{orange}{rgb}{0.87539215686858873,0.50482698971149964,0.12774509808012069}
\definecolor{yellow}{rgb}{0.8939061906626995,0.88929258034948033,0.29840446252594988}
\definecolor{pink}{rgb}{0.90225298544939825,0.56472128699807567,0.74206845059114346}

\usepackage{array}
\newcolumntype{L}[1]{>{\raggedright\let\newline\\\arraybackslash\hspace{0pt}}m{#1}}
\newcolumntype{C}[1]{>{\centering\let\newline\\\arraybackslash\hspace{0pt}}m{#1}}
\newcolumntype{R}[1]{>{\raggedleft\let\newline\\\arraybackslash\hspace{0pt}}m{#1}}
\urldef{\mail}\path|{p.moeskops}@tue.nl|

\newcommand{\keywords}[1]{\par\addvspace\baselineskip
\noindent\keywordname\enspace\ignorespaces#1}

\begin{document}

\mainmatter  

\title{Adversarial training and dilated convolutions for brain MRI segmentation}

\titlerunning{Adversarial training and dilated convolutions for brain MRI segmentation}
%
%
\author{Pim Moeskops\inst{1} \and Mitko Veta\inst{1} \and Maxime W. Lafarge\inst{1} \and Koen A.J. Eppenhof\inst{1} \and Josien P.W. Pluim\inst{1}}

\authorrunning{P. Moeskops et al.}


\institute{Medical Image Analysis Group, Department of Biomedical Engineering,\\ Eindhoven University of Technology, The Netherlands}

%
%

\maketitle

\begin{abstract}
Convolutional neural networks (CNNs) have been applied to various automatic image segmentation tasks in medical image analysis, including brain MRI segmentation. Generative adversarial networks have recently gained popularity because of their power in generating images that are difficult to distinguish from real images. 

In this study we use an adversarial training approach to improve CNN-based brain MRI segmentation. To this end, we include an additional loss function that motivates the network to generate segmentations that are difficult to distinguish from manual segmentations. During training, this loss function is optimised together with the conventional average per-voxel cross entropy loss.

The results show improved segmentation performance using this adversarial training procedure for segmentation of two different sets of images and using two different network architectures, both visually and in terms of Dice coefficients. 

\end{abstract}

\section{Introduction}
Convolutional neural networks (CNNs) have become a very popular method for medical image segmentation. In the field of brain MRI segmentation, CNNs have been applied to tissue segmentation \cite{Zhan15,Moes16,Moes16a} and various brain abnormality segmentation tasks \cite{Hava17,Kamn16,Ghaf17}. 

A relatively new approach for segmentation with CNNs is the use of dilated convolutions, where the weights of convolutional layers are sparsely distributed over a larger receptive field without losing coverage on the input image \cite{Yu16,Wolt16}. Dilated CNNs are therefore an effective approach to achieve a large receptive field with a limited number of trainable weights and a limited number of convolutional layers, without the use of subsampling layers. 

Generative adversarial networks (GANs) provide a method to generate images that are difficult to distinguish from real images \cite{Good14,Radf16,Wolt17}. To this end, GANs use a discriminator network that is optimised to discriminate real from generated images, which motivates the generator network to generate images that look real. A similar adversarial training approach has been used for domain adaptation, using a discriminator network that is trained to distinguish images from different domains \cite{Gani16,Kamn16a} and for improving image segmentations, using a discriminator network that is trained to distinguish manual from generated segmentations \cite{Luc16}. Recently, such a segmentation approach has also been applied in medical imaging for the segmentation of prostate cancer in MRI \cite{Kohl17} and organs in chest X-rays \cite{Dai17}.

In this paper we employ adversarial training to improve the performance of brain MRI segmentation in two sets of images using a fully convolutional and a dilated network architecture. 

\section{Materials and Methods}
\subsection{Data}
\subsubsection{Adult subjects}
35 T\textsubscript{1}-weighted MR brain images (15 training, 20 test) were acquired on a Siemens Vision 1.5T scanner at an age ($\mu\pm\sigma$) of 32.9 $\pm$ 19.2 years, as provided by the MICCAI 2012 challenge on multi-atlas labelling \cite{Land12}. The images were segmented in six classes: white matter (WM), cortical grey matter (cGM), basal ganglia and thalami (BGT), cerebellum (CB), brain stem (BS), and lateral ventricular cerebrospinal fluid (lvCSF). 

\subsubsection{Elderly subjects}
20 axial T\textsubscript{1}-weighted MR brain images (5 training, 15 test) were acquired on a Philips Achieva 3T scanner at an age ($\mu\pm\sigma$) of 70.5 $\pm$ 4.0 years, as provided by the MRBrainS13 challenge \cite{Mend15etal}. The images were segmented in seven classes: WM, cGM, BGT, CB, BS, lvCSF, and peripheral cerebrospinal fluid (pCSF). Possible white matter lesions were included in the WM class. 

\subsection{Network architecture}
Two different network architectures are used to evaluate the hypothesis that adversarial training can aid in improving segmentation performance: a fully convolutional network and a network with dilated convolutions. The outputs of these networks are input for a discriminator network, which distinguishes between generated and manual segmentations. The fully convolutional nature of both networks allows arbitrarily sized inputs during testing. Details of both segmentation networks are listed in Figure \ref{fig:overview}, left. 

\subsubsection{Fully convolutional network}
A network with 15 convolutional layers of 32 3$\times$3 kernels is used (Figure \ref{fig:overview}, left), which results in a receptive field of 31$\times$31 voxels. During training, an input of 51$\times$51 voxels is used, corresponding to an output of 21$\times$21 voxels. The network has 140,039 trainable parameters for $C=7$ classes (6 plus background; adult subjects) and 140,296 trainable parameters for $C=8$ classes (7 plus background; elderly subjects). 

\subsubsection{Dilated network}
\begin{figure}
\centering

\begin{subfigure}{.4\textwidth}
\centering
\scriptsize
\begin{tabular}[htb]{c c c c}
\hline
\multicolumn{4}{|c|}{\textbf{Fully convolutional network}} \\
\hline
\multicolumn{1}{|c}{Kernel size} & \multicolumn{1}{|c}{Dilation} & \multicolumn{1}{|c}{Kernels} & \multicolumn{1}{|c|}{Layers}\\
\hline
\multicolumn{1}{|c}{3$\times$3} & 1 & 32 &  \multicolumn{1}{c|}{15}\\ 
\multicolumn{1}{|c}{1$\times$1} & 1 & 256 &  \multicolumn{1}{c|}{1}\\
\multicolumn{1}{|c}{1$\times$1} & 1 & $C$ &  \multicolumn{1}{c|}{1}\\ 
\hline
&&&\\
\hline
\multicolumn{4}{|c|}{\textbf{Dilated network}}\\\hline
\multicolumn{1}{|c}{Kernel size} & \multicolumn{1}{|c}{Dilation} & \multicolumn{1}{|c}{Kernels} & \multicolumn{1}{|c|}{Layers}\\\hline
\multicolumn{1}{|c}{3$\times$3} & 1 & 32 & \multicolumn{1}{c|}{2}\\
\multicolumn{1}{|c}{3$\times$3} & 2 & 32 & \multicolumn{1}{c|}{1}\\
\multicolumn{1}{|c}{3$\times$3} & 4 & 32 & \multicolumn{1}{c|}{1}\\
\multicolumn{1}{|c}{3$\times$3} & 8 & 32 & \multicolumn{1}{c|}{1}\\
\multicolumn{1}{|c}{3$\times$3} & 16 & 32 & \multicolumn{1}{c|}{1}\\
\multicolumn{1}{|c}{3$\times$3} & 1 & 32 & \multicolumn{1}{c|}{1}\\
\multicolumn{1}{|c}{1$\times$1} & 1 & $C$ & \multicolumn{1}{c|}{1}\\%
\hline
\end{tabular}
\caption{Segmentation networks}
\end{subfigure}
\hspace{15mm}
\begin{subfigure}{.44\textwidth}
\centering
\includegraphics[trim=0mm 0mm 0mm 0mm,clip,width=\textwidth]{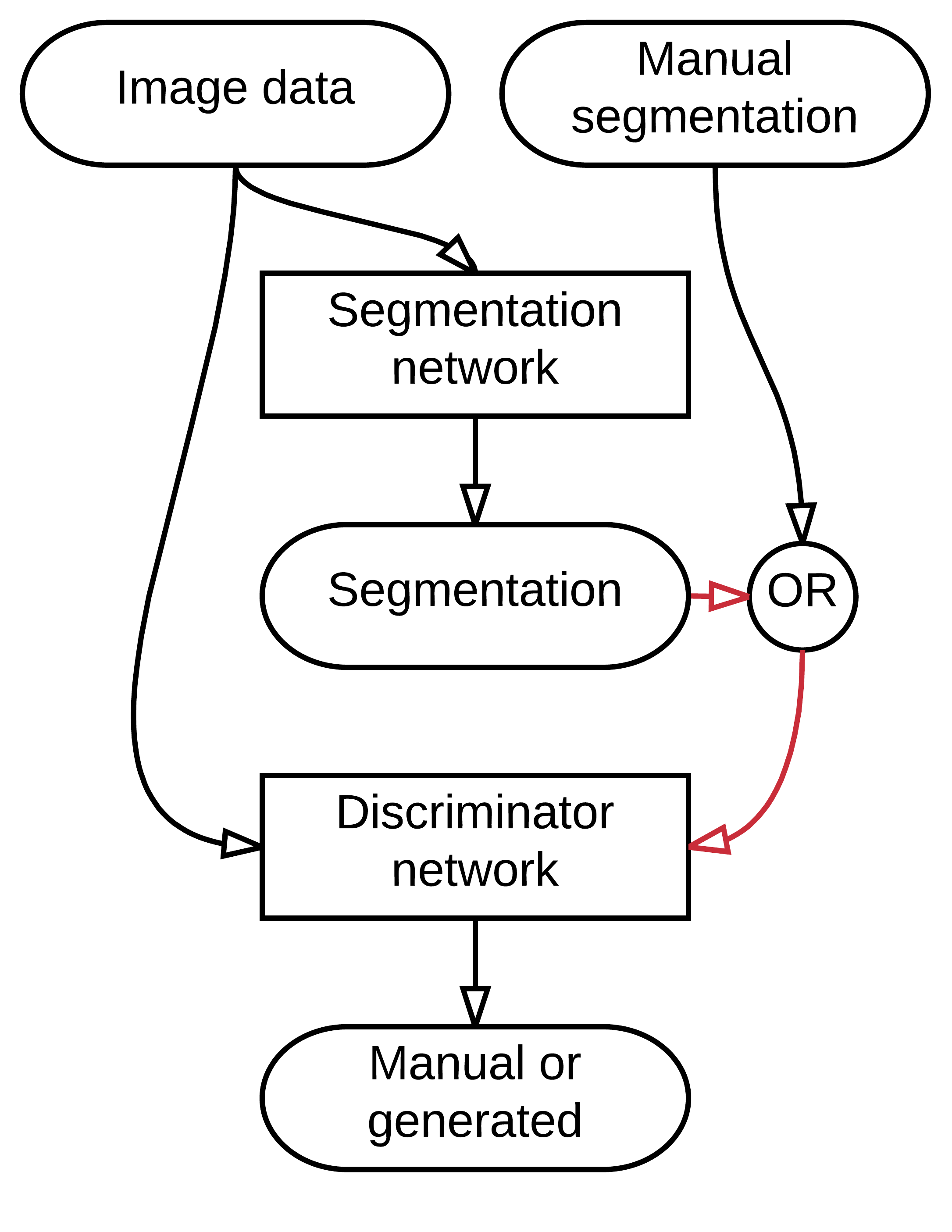}
\caption{Adversarial training}
\end{subfigure}
\caption{Left: Segmentation network architectures for the 17-layer fully convolutional (top) and 8-layer dilated (bottom) segmentation networks. The receptive fields are 67$\times$67 for the dilated network and 31$\times$31 for the fully convolutional network. No subsampling layers are used in both networks. Right: Overview of the adversarial training procedure. The red connections indicate how the discriminator loss influences the segmentation network during backpropagation.} \label{fig:overview}
\end{figure}

The dilated network uses the same architecture as proposed by Yu et al. \cite{Yu16}, which uses layers of 3$\times$3 kernels with increasing dilation factors (Figure \ref{fig:overview}, left). This results in a receptive field of 67$\times$67 voxels using only 7 layers of 3$\times$3 convolutions, without any subsampling layers. During training, an input of 87$\times$87 voxels is used, which corresponds to an output of 21$\times$21 voxels. In each layer 32 kernels are trained. The network has 56,039 trainable parameters for $C=7$ classes (6 plus background; adult subjects) and 56,072 trainable parameters for $C=8$ classes (7 plus background; elderly subjects). 

\subsubsection{Discriminator network}
The input to the discriminator network are the segmentation, as one-hot encoding or softmax output, and image data in the form of a 25$\times$25 patch. In this way, the network can distinguish real from generated combinations of image and segmentation patches. The image patch and the segmentation are concatenated after two layers of 3$\times$3 kernels on the image patch. The discriminator network further consists of three layers of 32 3$\times$3 kernels, a 3$\times$3 max-pooling layer, two layers of 32 3$\times$3 kernels, and a fully connected layer of 256 nodes. The output layer with two nodes, distinguishes between manual and generated segmentations. 

\begin{figure}[htb]
\begin{center}
\begin{subfigure}{.31\textwidth}
\vspace{-3mm}
\caption*{Adult subject (FCN)}
\vspace{-1mm}
\includegraphics[trim=10mm 20mm 20mm 10mm,clip,width=\textwidth]{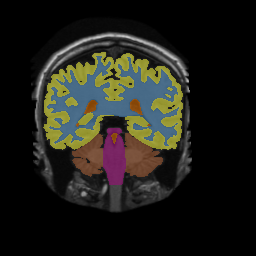}\\
\vspace{-3mm}
\caption*{Adult subject (DN)}
\vspace{-1mm}
\includegraphics[trim=10mm 30mm 20mm 5mm,clip,width=\textwidth]{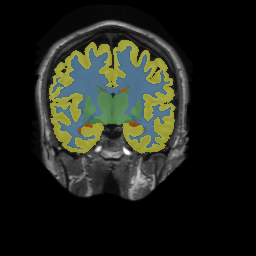}\\
\vspace{-3mm}
\caption*{Elderly subject (FCN)}
\vspace{-1mm}
\includegraphics[trim=10mm 1mm 10mm 5mm,clip,width=\textwidth]{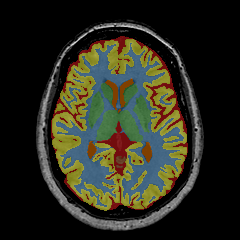}\\
\vspace{-3mm}
\caption*{Elderly subject (DN)}
\vspace{-1mm}
\includegraphics[trim=10mm 7mm 10mm 4mm,clip,width=\textwidth]{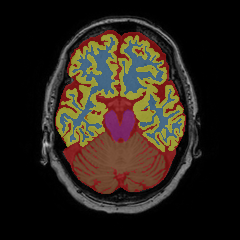}
\caption{Reference} 

\end{subfigure}
\begin{subfigure}{.31\textwidth}
\vspace{-3mm}
\caption*{}
\vspace{-1mm}
\includegraphics[trim=40mm 80mm 80mm 40mm,clip,width=\textwidth]{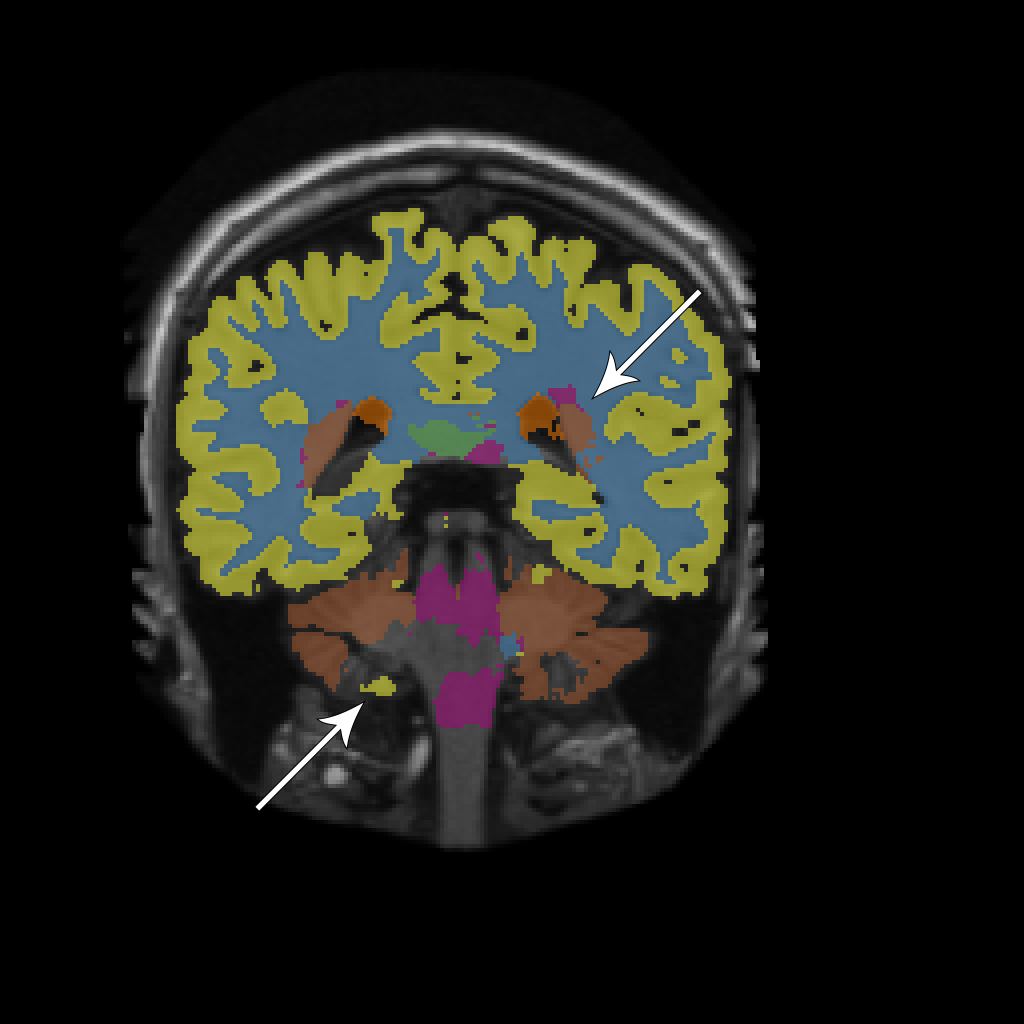}\\
\vspace{-3mm}
\caption*{}
\vspace{-1mm}
\includegraphics[trim=40mm 120mm 80mm 20mm,clip,width=\textwidth]{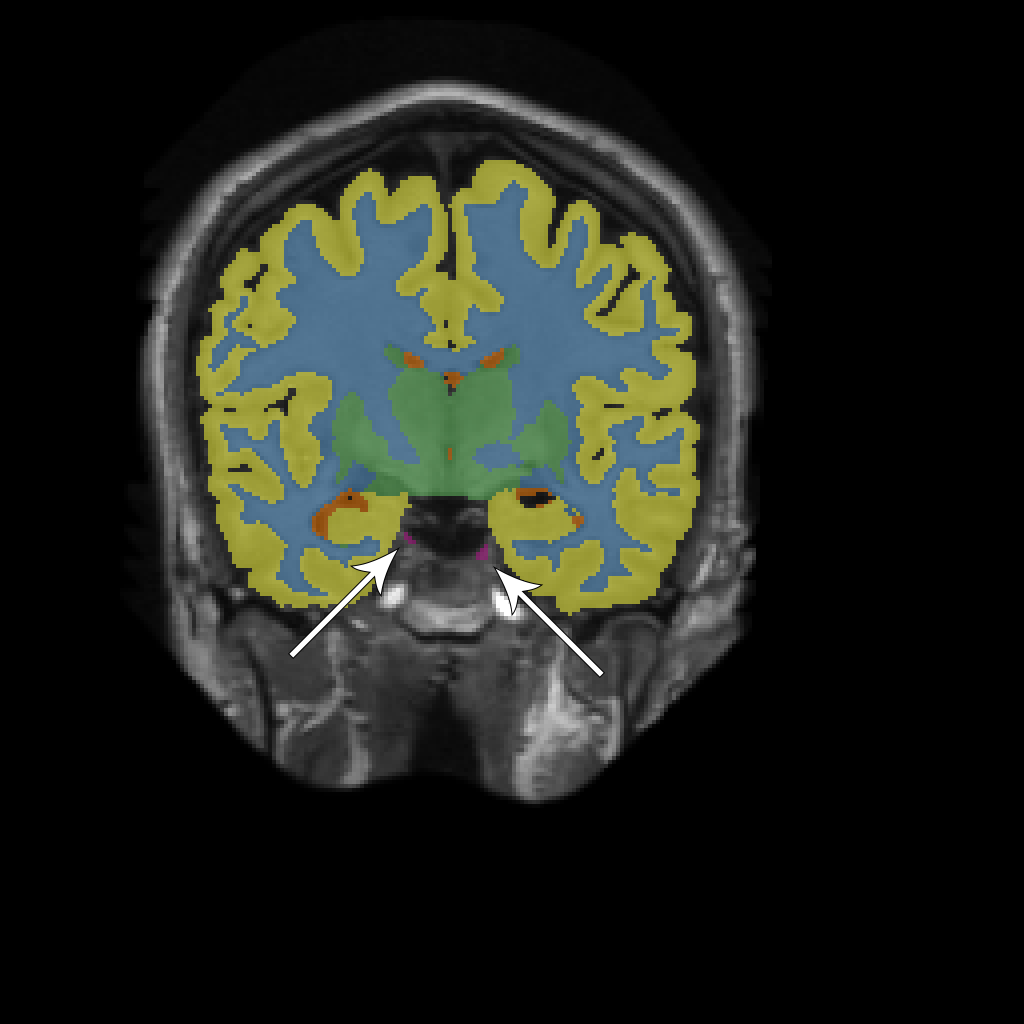}\\
\vspace{-3mm}
\caption*{}
\vspace{-1mm}
\includegraphics[trim=40mm 4mm 40mm 20mm,clip,width=\textwidth]{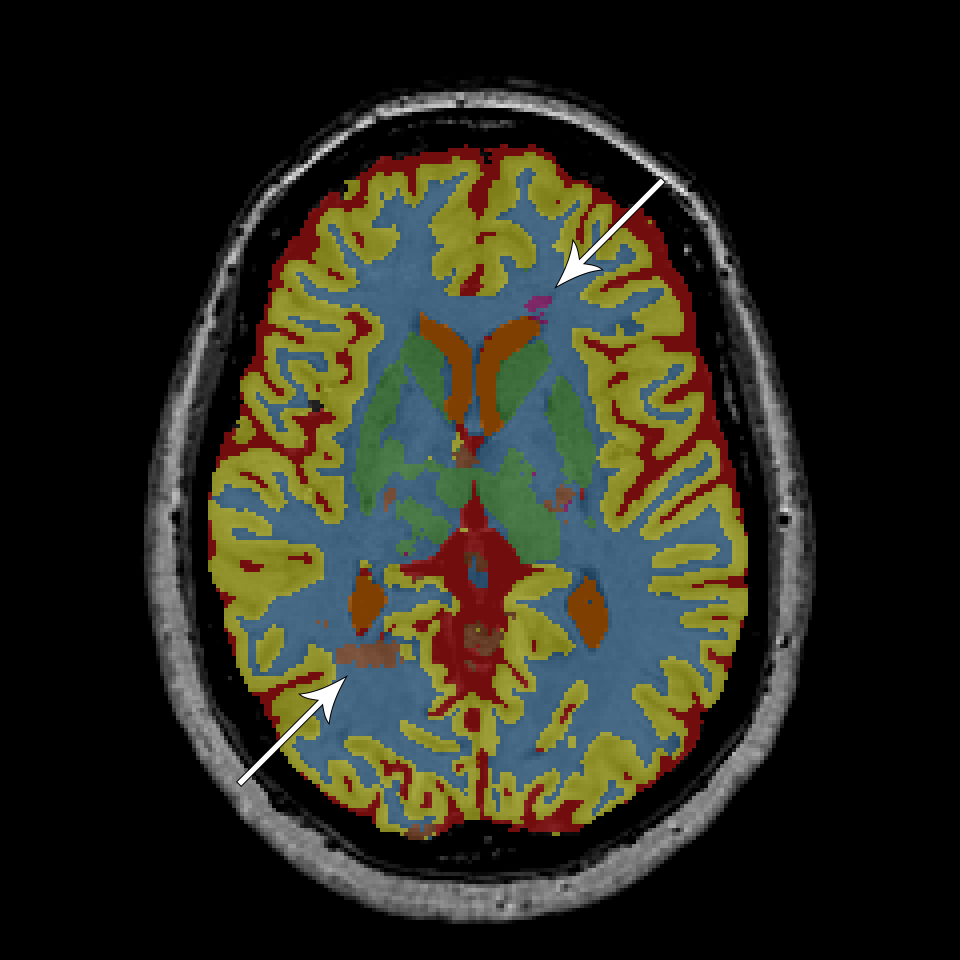}\\
\vspace{-3mm}
\caption*{}
\vspace{-1mm}
\includegraphics[trim=40mm 28mm 40mm 16mm,clip,width=\textwidth]{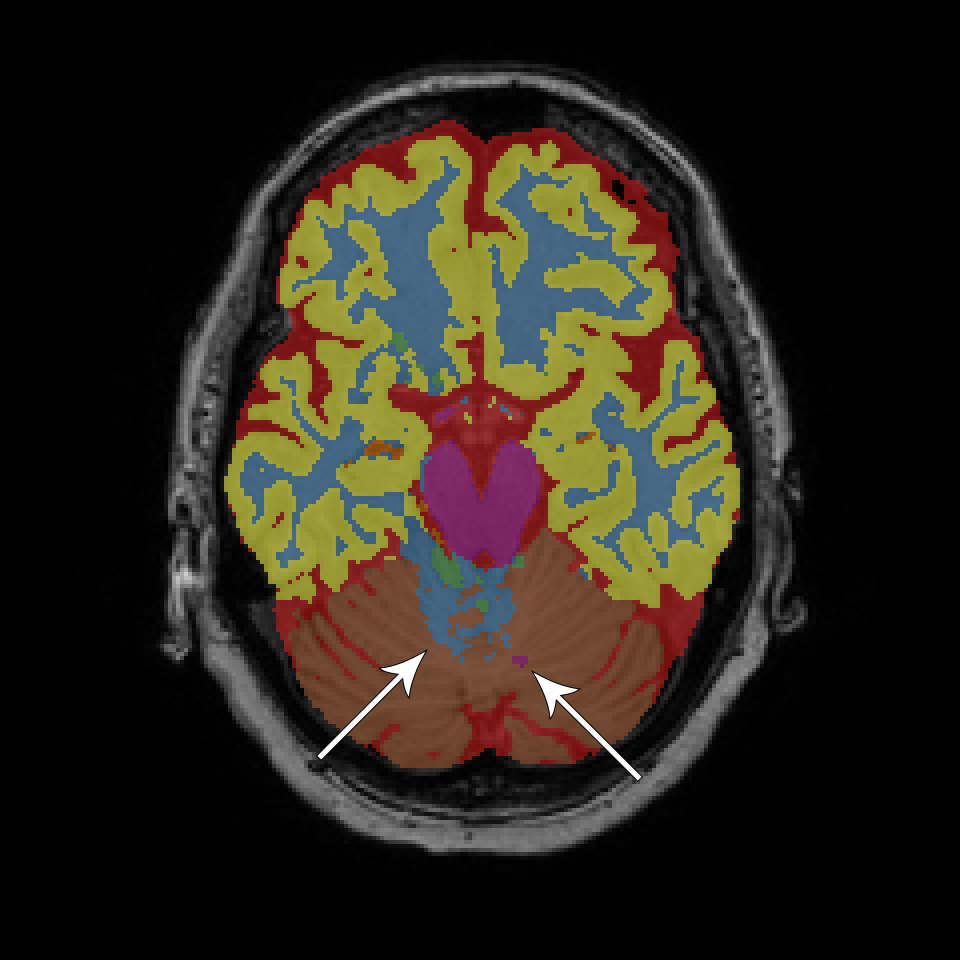}
\caption{Without adversarial} 
\end{subfigure}
\begin{subfigure}{.31\textwidth}
\vspace{-3mm}
\caption*{}
\vspace{-1mm}
\includegraphics[trim=10mm 20mm 20mm 10mm,clip,width=\textwidth]{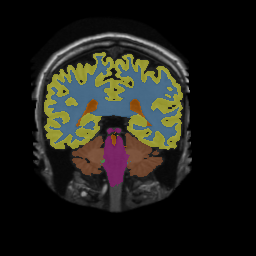}\\
\vspace{-3mm}
\caption*{}
\vspace{-1mm}
\includegraphics[trim=10mm 30mm 20mm 5mm,clip,width=\textwidth]{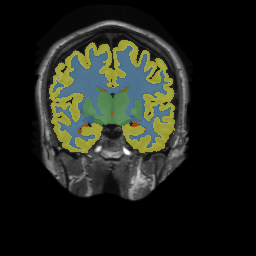}\\
\vspace{-3mm}
\caption*{}
\vspace{-1mm}
\includegraphics[trim=10mm 1mm 10mm 5mm,clip,width=\textwidth]{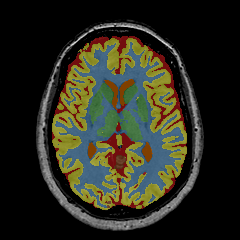}\\
\vspace{-3mm}
\caption*{}
\vspace{-1mm}
\includegraphics[trim=10mm 7mm 10mm 4mm,clip,width=\textwidth]{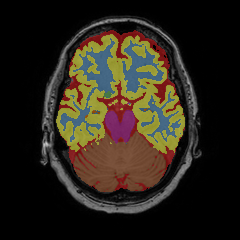}
\caption{With adversarial} 
\end{subfigure}
\end{center}
\vspace{-7mm}
\caption{Example segmentation results in four of the test images. From top to bottom: an adult subject using the fully convolutional network (FCN), an adult subject using the dilated network (DN), an elderly subject using the fully convolutional network (FCN), and an elderly subject using the dilated network (DN). The colours are as follows: WM in blue, cGM in yellow, BGT in green, CB in brown, BS in purple, lvCSF in orange, and pCSF in red. The arrows indicate errors that were corrected when the adversarial training procedure was used.} \label{fig:seg}
\end{figure}

\FloatBarrier

\subsection{Adversarial training}
An overview of the adversarial training procedure is shown in Figure \ref{fig:overview}, right.

Three types of updates for the segmentation network parameters $\theta_s$ and the discriminator network parameters $\theta_d$ are possible during the training procedure: (1) an update of only the segmentation network based on the cross-entropy loss over the segmentation map, $L_s(\theta_s)$, (2) an update of the discriminator network based on the discrimination loss using a manual segmentation as input, $L_d(\theta_d)$, and (3) an update of the whole network (segmentation and discriminator network) based on the discriminator loss using an image as input, $L_a(\theta_s,\theta_d)$. Only $L_s(\theta_s)$ and $L_a(\theta_s,\theta_d)$ affect the segmentation network. The parameters $\theta_s$ are updated to maximise the discriminator loss $L_a(\theta_s,\theta_d)$, i.e. the updates for the segmentation network are performed in the direction to ascend the loss instead of to descend the loss.

The three types of updates are performed in an alternating fashion. The updates based on the segmentation loss and the updates based on the discriminator loss are performed with separate optimisers using separate learning rates. Using a smaller learning rate, the discriminator network adapts more slowly than the segmentation network, such that the discriminator loss does not converge too quickly and can have enough influence on the segmentation network. 

For each network, rectified linear units are used throughout, batch normalisation \cite{Ioff15} is used on all layers and dropout \cite{Sriv14} is used for the 1$\times$1 convolution layers.

\begin{figure}[htb]
\begin{center}
\includegraphics[trim=5mm 0mm 20mm 0mm,clip,width=0.49\textwidth]{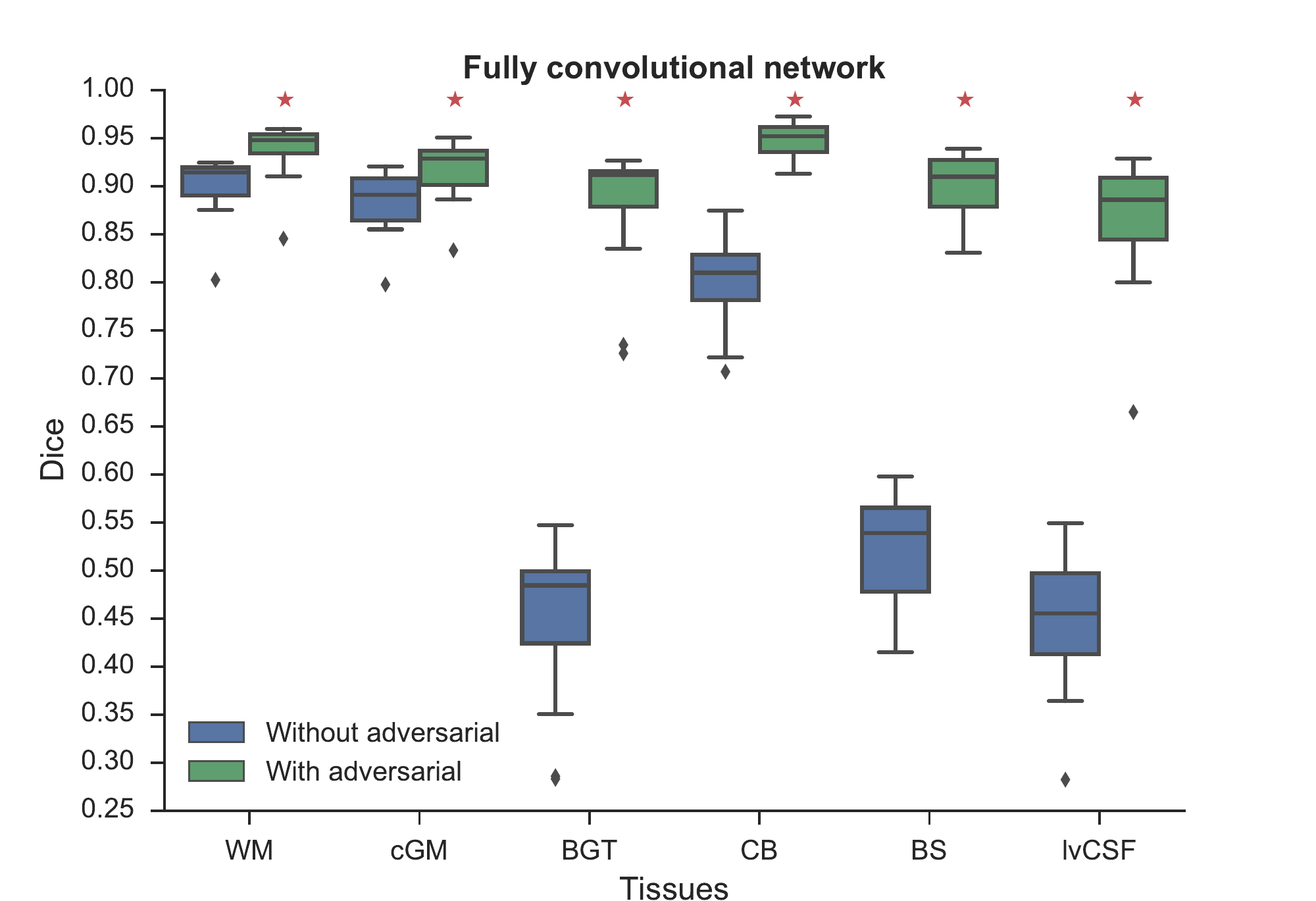}
\includegraphics[trim=5mm 0mm 20mm 0mm,clip,width=0.49\textwidth]{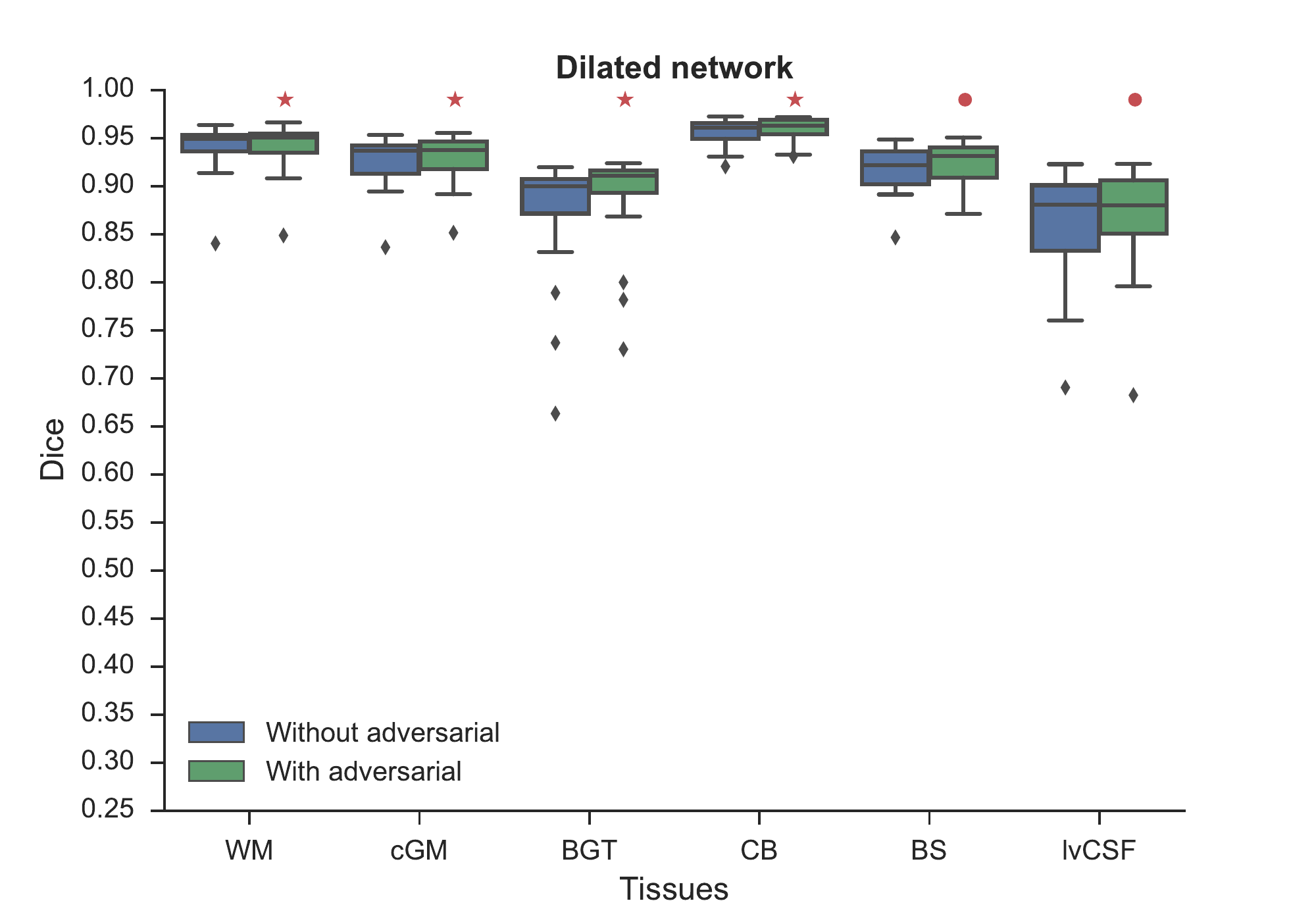}
\includegraphics[trim=5mm 0mm 20mm 0mm,clip,width=0.49\textwidth]{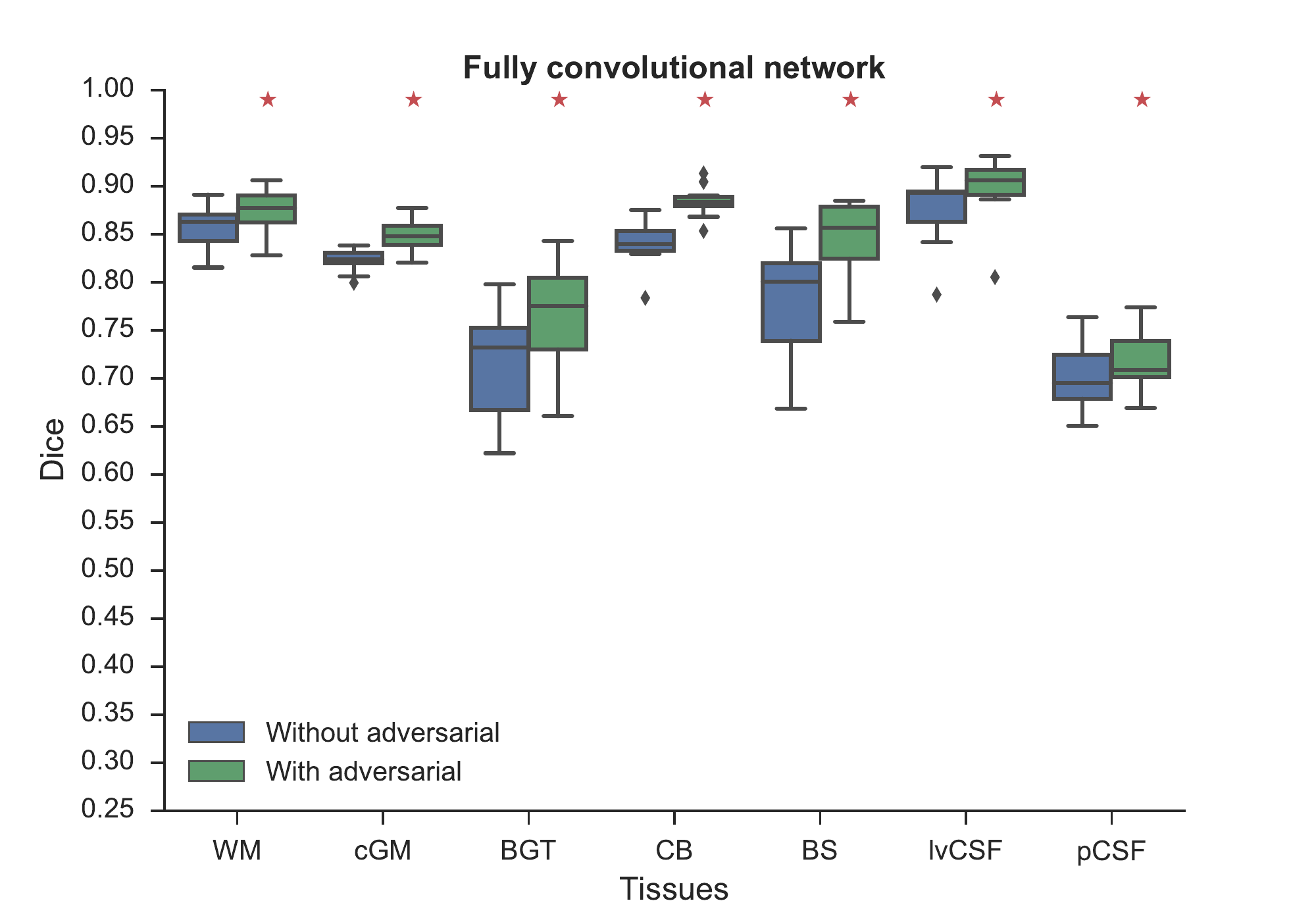}
\includegraphics[trim=5mm 0mm 20mm 0mm,clip,width=0.49\textwidth]{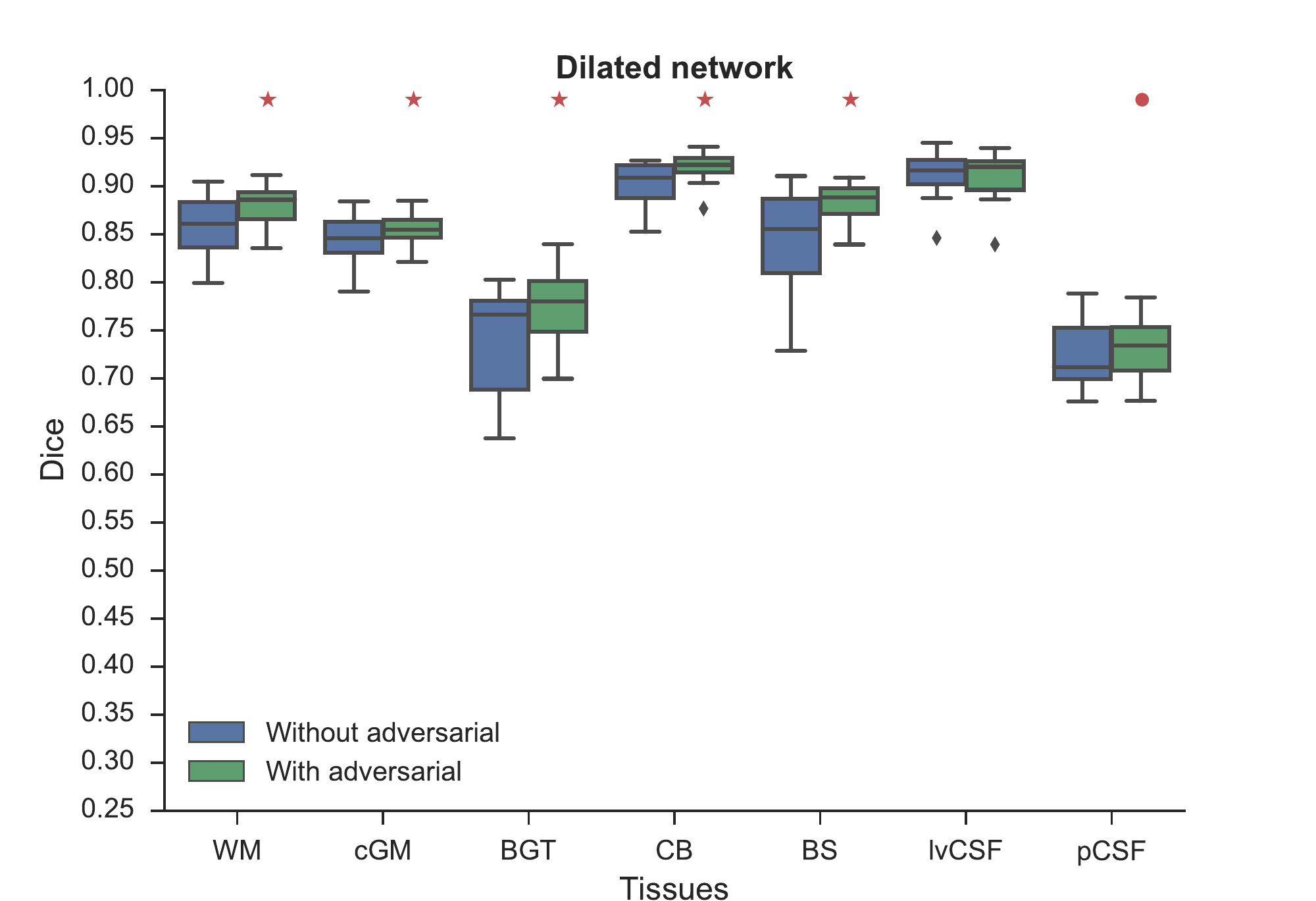}
\end{center}
\caption{Dice coefficients for the adult subjects (top row) and the elderly subjects (bottom row) for white matter (WM), cortical grey matter (cGM), basal ganglia and thalami (BGT), cerebellum (CB), brain stem (BS), lateral ventricular cerebrospinal fluid (lvCSF), and peripheral cerebrospinal fluid (pCSF), without (blue) and with (green) adversarial training. Left column: fully convolutional network. Right column: dilated network. Red stars ($p<0.01$) and red circles ($p<0.05$) indicate significant improvement based on paired $t$-tests.} \label{fig:Dice}
\end{figure}

\section{Experiments and Results}
\subsection{Experiments}
As a baseline, the segmentation networks are trained without the adversarial network. The updates are performed with RMSprop using a learning rate of $10^{-3}$ and minibatches of 300 samples. The networks are trained in 5 epochs, where each epoch corresponds to 50,000 training patches per class per image. Note that during this training sample balancing process, the class label corresponds to the label of the central voxel, even though a larger image patch is labelled.

The discriminator and segmentation network are trained using the alternating update scheme. The updates for both loss functions are performed with RMSprop using a learning rate of $10^{-3}$ for the segmentation loss and a learning rate of $10^{-5}$ for the discriminator loss. The updates alternate between the $L_s$, $L_d$ and $L_a$ loss functions, using minibatches of $300/3=100$ samples for each. 

\subsection{Evaluation}
Figure \ref{fig:seg} provides a visual comparison between the segmentations obtained with and without adversarial training, showing that the adversarial approach generally resulted in less noisy segmentations. The same can be seen from the total number of 3D components (including the background class) that compose the segmentations. For the adult subjects, the number of components per image ($\mu\pm\sigma$) decreased from $1745\pm400$ to $626\pm247$ using the fully convolutional network and from $417\pm152$ to $365\pm122$ using the dilated network. For the elderly subjects, the number of components per image ($\mu\pm\sigma$) decreased from $926\pm134$ to $692\pm88$ using the fully convolutional network and from $601\pm104$ to $481\pm90$ using the dilated network. 

The evaluation results in terms of Dice coefficients (DC) between the automatic and manual segmentations are shown in Figure \ref{fig:Dice} as boxplots. Significantly improved DC, based on paired $t$-tests, were obtained for each of the tissue classes, in both image sets, and for both networks. The only exception was lvCSF in the elderly subjects using the dilated network. For the adult subjects, the DC averaged over all 6 classes ($\mu\pm\sigma$) increased from $0.67\pm0.04$ to $0.91\pm0.03$ using the fully convolutional network and from $0.91\pm0.03$ to $0.92\pm0.03$ using the dilated network. For the elderly subjects, the DC averaged over all 7 classes ($\mu\pm\sigma$) increased from $0.80\pm0.02$ to $0.83\pm0.02$  using the fully convolutional network and from $0.83\pm0.02$ to $0.85\pm0.01$ using the dilated network.

\section{Discussion and Conclusions}
We have presented an approach to improve brain MRI segmentation by adversarial training. The results showed improved segmentation performance both qualitatively (Figure \ref{fig:seg}) and quantitatively in terms of DC (Figure \ref{fig:Dice}). The improvements were especially clear for the deeper, more difficult to train, fully convolutional networks as compared with the more shallow dilated networks. Furthermore, the approach improved structural consistency, e.g. visible from the reduced number of components in the segmentations. Because these improvements were usually small in size, their effect on the DC was limited.

The approach includes an additional loss function that distinguishes between real and generated segmentations and can therefore capture inconsistencies that a normal per-voxel loss averaged over the output does not capture. The proposed approach can be applied to any network architecture that, during training, uses an output in the form of an image patch, image slice, or full image instead of a single pixel/voxel. 

Various changes to the segmentation network that might improve the results could be evaluated in future work, such as different receptive fields, multiple inputs, skip-connections, 3D inputs, etc. Using a larger output patch size or even the whole image as output could possibly increase the effect of the adversarial training by including more information that could help in distinguishing manual from generated segmentations. This could, however, also reduce the influence of local information, resulting in a too global decision. Further investigation is necessary to evaluate which of the choices in the network architecture and training procedure have most effect on the results.

\vspace{3mm}
\textbf{Acknowledgements} 
The authors would like to thank the organisers of MRBrainS13 and the multi-atlas labelling challenge for providing the data. The authors gratefully acknowledge the support of NVIDIA Corporation with the donation of a Titan X Pascal GPU.

\FloatBarrier

\bibliographystyle{splncs03}
\bibliography{CAD}

\end{document}